\pdfoutput=1

\documentclass[11pt]{article}
\usepackage{authblk}
\usepackage{ACL2023}
\usepackage{mathtools}
\usepackage{amssymb}
\usepackage{cleveref}
\usepackage{graphicx}
\usepackage{booktabs}
\usepackage{amsmath}
\usepackage{times}
\usepackage{latexsym}

\usepackage{dsfont}
\usepackage{bm}

\usepackage[T1]{fontenc}

\usepackage[utf8]{inputenc}

\usepackage{microtype}

\usepackage{inconsolata}

\title{Learning from Response not Preference: A Stackelberg Approach for LLM Detoxification using Non-parallel Data }

\author{
\textbf{Xinhong Xie\textsuperscript{1}},
\textbf{Tao Li\textsuperscript{2}},
\textbf{Quanyan Zhu\textsuperscript{2}}
\\
\textsuperscript{1}Department of Computer Science and Engineering, The Pennsylvania State University,\\
\textsuperscript{2}Department of Electrical and Computer Engineering, New York University
\\
 \small{
   \textbf{Correspondence:} \href{mailto:taoli@nyu.edu}{taoli@nyu.edu}
 }
}

\begin{document}
\maketitle

\begin{abstract}
Text detoxification, a variant of style transfer tasks, finds useful applications in online social media. This work presents a fine-tuning method that only uses non-parallel data to turn large language models (LLM) into a detoxification rewritter. We model the fine-tuning process as a Stackelberg game between an LLM (leader) and a toxicity screener (follower), which is a binary style classifier (toxic or non-toxic). The LLM aims to align its preference according to the screener and generate paraphases passing the screening. The primary challenge of non-parallel data fine-tuning is incomplete preference. In the case of unsuccessful paraphrases, the classifier cannot establish a preference between the input and paraphrase, as they belong to the same toxic style. Hence, preference-alignment fine-tuning methods, such as direct preference optimization (DPO), no longer apply. To address the challenge of incomplete preference, we propose Stackelberg response optimization (SRO), adapted from DPO, to enable the LLM to learn from the follower's response. The gist is that SRO decreases the likelihood of generating the paraphrase if it fails the follower's screening while performing DPO on the pair of the toxic input and its paraphrase when the latter passes the screening. Experiments indicate that the SRO-fine-tunned LLM achieves satisfying performance comparable to state-of-the-art models regarding style accuracy, content similarity, and fluency. The overall detoxification performance surpasses other computing methods and matches the human reference. Additional empirical evidence suggests that SRO is sensitive to the screener's feedback, and a slight perturbation leads to a significant performance drop. We
release the code and LLM models at \url{https://github.com/XXXinhong/Detoxification_LLM}.      
\end{abstract}

\section{Introduction}
Identification of toxicity and other undesirable contents in user-generated texts is an active research area in NLP~\cite{pan24overview}. As a proactive combat (besides deletion), the task of automatic rewriting/rephrasing has received increasing attention from the NLP community \cite{VillateCastillo2024Toxicity}.  

Most existing works on text detoxification casts the problem as a variant of \textit{style transfer}: the task of changing the style of a text while preserving the content. The style of a text refers to its characteristics, such as sentiment, level of formality, and political inclination \cite{dale-etal-2021-text}. As a sequence-to-sequence task, style transfer tasks can employ an encoder-decoder model trained on parallel data that includes a parallel corpus of toxic sentences and their non-toxic paraphrases \cite{wieting-gimpel-2018-paranmt}.  Despite recent efforts to build parallel datasets \cite{logacheva-etal-2022-paradetox}, the existing parallel data collection is still in its infancy. As an alternative, style transfer models can also utilize non-parallel data. Prior works train encoder-decoder models on non-parallel data and push the decoder toward the target style with the help of style classifiers \cite{lee-2020-stable}.  

\begin{figure}
    \centering
\includegraphics[width=1\linewidth]{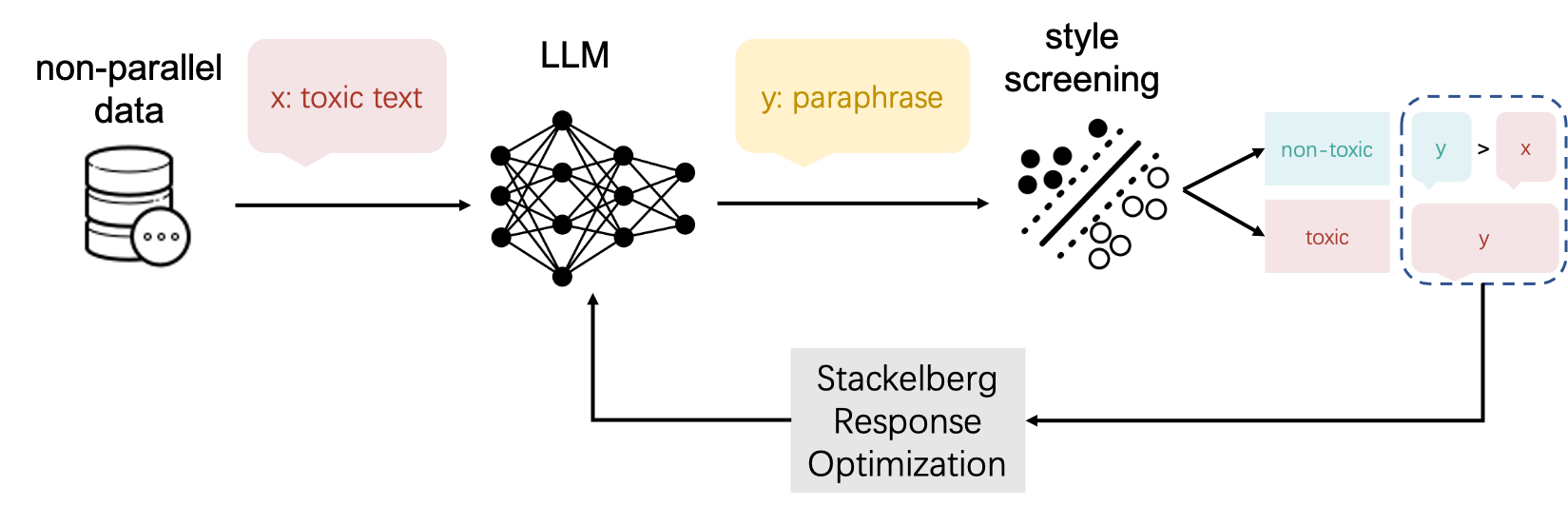}
    \caption{An illustration of Stackelberg Response Optimization (SRO), which addresses incomplete preference in detoxification tasks. For a toxic input $x$, when the paraphrase $y$ passes the screening, SRO performs DPO on the pair $(x,y)$; otherwise, it decreases the likelihood of generating $y$.    }
    \label{fig:sro}
\end{figure}
This work considers LLM-based detoxification models trained on non-parallel data. Without parallel data, few-shot fine-tuning does not apply since concrete demonstrations are not available to LLMs. Hence, we consider a different perspective on text detoxification: casting the problem as an alignment problem, where an LLM needs to be fine-tuned to align with a given preference (e.g., non-toxic over toxic contents). Such preference alignment has gained soaring popularity in LLM research as it addresses the LLM's steerability issue, aiming to achieve precise control of LLM's behaviors. Such fine-tuning requires human preference labeling, which is instrumental in reinforcement learning from human feedback (RLHF) pipelines \cite{ouyang2022traininglanguagemodelsfollow}. However, such a preference relation may not be immediately available in detoxification. Given a toxic text input, when the LLM generates two outputs that are still toxic, it is unclear which one is less toxic since both belong to the same style. In practice, we also observe that the human-labeled non-parallel data only involves binary labels (toxic or non-toxic) without explicit preference relations, making it challenging to rank texts from the same class. Consequently, RLHF may not apply due to the lack of clear preferences.   

From the example above, the key challenge of applying LLMs to detoxification (or style transfer in general) from the alignment viewpoint is that \textit{no preference exists among texts belonging to the same style}. Therefore, LLMs need to learn from the outcome of style transfer, i.e., the feedback from the style screening (either success or failure), rather than preference labeling. 

We model the LLM's fine-tuning process as a Stacelberg game between the LLM and a style classifier trained from human-labeled data. As illustrated in \Cref{fig:sro}, when the LLM (leader) receives a toxic text from the training dataset, it first generates a paraphrase and sends it to the screener (follower), who responds with a binary outcome: either success or failure.  Based on this response, LLM updates the model weights by applying the gradient to our proposed Stackelberg loss. The intuition behind such loss is that 1) if the detoxification is successful, then the pair of the input text and LLM-generated text presents a clear preference, which is perfectly handled by the original DPO pipeline, and the Stackelberg loss reduces to DPO loss; 2) otherwise, only the failed transfer is utilized to compute the loss that aims to penalize LLMs for such paraphrasing. The sheer difference between our proposed fine-tuning method and RLHF is that the LLM learns from the classifier's response to its generated text without requiring additional preference labeling. We refer to the proposed fine-tuning method as Stackelberg Response Optimization. \textbf{Our contributions} are as follows.
\begin{itemize}
    \item We model LLM fine-tuning for detoxification as a Stacklberg game between the LLM and style screener. Such a Stackelberg formulation spares one from parallel data and human labeling. 
    \item We propose Stackelberg Response Optimization (SRO), adapted from DPO, to address the issue of missing preferences in RLHF.  
    \item We conduct an evaluation of the fine-tuned LLM and its comparison with a number of state-of-the-art models in text detoxification tasks. Our proposed STALR achieves superior overall performance (a joint metric synthesizing style accuracy, content similarity, and fluency) over other computing models and matches human reference.   
    \item We examine SRO's robustness to the screener. Empirical evidence suggests that SRO is highly sensitive to the screener's feedback. A slight perturbation in screening significantly degrades the detoxification performance.
\end{itemize}

\section{Related Works}
\paragraph{Style transfer and detoxification}
Prior works in style transfer consider encoder-decoder models on non-parallel data while using adversarial classifiers to steer the decoder toward the desired style \cite{10.5555/3295222.3295427}. To address the challenge of lacking parallel data, researchers have employed methods such as reinforcement learning \cite{ijcai2019p711}, amortized variational inference \cite{He2020A}, and utilizing style transfer classifiers to train models that convert texts from one style to another and vice versa \cite{lee-2020-stable}. 

Detoxification specifically involves converting toxic to neutral language using datasets not directly paired but categorized for toxicity, treating toxic and neutral texts as distinct groups (nonparallel data). Several studies use datasets like the Jigsaw series \cite{laugier-etal-2021-civil} or create their own from social media platforms for this purpose \cite{jigsaw-multilingual-toxic-comment-classification, jigsaw-toxic-comment-classification-challenge, jigsaw-unintended-bias-in-toxicity-classification}. 

Prior detoxification models often borrow methodologies from other domains \cite{shen2017styletransfernonparalleltext} and \cite{Fu_Tan_Peng_Zhao_Yan_2018}, using autoencoders \cite{nogueira-dos-santos-etal-2018-fighting} with style classification and cycle-consistency losses, or fine-tuning denoising autoencoders like T5 \cite{laugier-etal-2021-civil}. Some models apply direct corrections to specific words \cite{tran-etal-2020-towards, ijcai2019p732} and then enhance text fluency with sequence-to-sequence models. Similarly, \cite{dale-etal-2021-text} uses masked language models for targeted edits. 

\paragraph{Preference alignment and fine-tuning of LLM}

LLMs have shown great capabilities in the past few years in generative tasks across various domains \cite{Bubeck2023SparksOA,li2024symbiotic}, and their encouraging success relies on the precise control of LLMs' output through alignment fine-tuning. One of the popular methods for gaining such control is reinforcement learning with human feedback \cite{ouyang2022traininglanguagemodelsfollow}. By training a reward model on human-labeled data, one can use the reward model to fine-tune LLM models using RL algorithms. 

To avoid bias that exists in the reward model, \cite{rafailov2024directpreferenceoptimizationlanguage} develops Direct Preference Optimization (DPO), which does not rely on another reward model to calculate loss; instead, DPO calculates the loss based on the distance between logits and the expected answers. In addition to DPO, there are also some emerging techniques, such as Identity Preference Optimisation (IPO)~\cite{azar2023generaltheoreticalparadigmunderstand}, which avoid Bradley-Terry modelization assumption to avoid overfitting problems, and Kahneman-Tversky Optimization (KTO)~\cite{ethayarajh2024ktomodelalignmentprospect}, which apply human-aware losses (HALOs) to directly maximizing the utility of generations instead of maximizing the likelihood of preferences.

\paragraph{Game-theoretic RLHF}

Multi-agent reinforcement learning offers a robust theoretical grounding for examining learning agents that interact and need to adjust their strategies in response \cite{tao22confluence,tao_info}. Recent developments in game-theoretic methods tailored to RLHF include Self-Play Preference Optimization (SPO) \cite{swamy2024minimaximalistapproachreinforcementlearning}, which utilizes self-play to compare win rates across different trajectories. Similarly, Self-Play fIne-tuNing (SPIN) \cite{chen2024selfplayfinetuningconvertsweak} employs self-play to create synthetic high-quality data by differentiating self-generated continuations from those generated by humans. Both approaches utilize a consistent reward model akin to traditional RLHF. Additionally, Nash-RLHF \cite{munos2024nashlearninghumanfeedback} introduces a competitive setup involving two language models, each aiming to gain favor from a human evaluator.

Closely related to our Stackelberg formulation, Chakraborty et al. \cite{chakraborty2024parlunifiedframeworkpolicy} also consider the preference alignment as a Stackelberg game, where the reward model acts as the leader and the RL agent as the follower. However, the reward model is still built on complete preference relations, whereas our work focuses on incomplete preference resulting from the binary feedback of the screener. 

\section{Preliminary}
\paragraph{Notations} Consider a non-parallel dataset denoted by $D$, which consists of a blend of toxic and non-toxic texts (sentences). We denote by $x\in D$ A typical text from the dataset. In our text detoxification task, such a text $x$ is often paired with a prompt instructing the LLM to rewrite the sentence. For example, the prompt used in our experiment is ``paraphrase the following: $x$.'' LLM, despite its intricate inner workings, is represented by a function $f_\theta(\cdot)$ that maps the input text $x$ to a rephrased text $y$, where $\theta\in \Theta$ denotes the LLM's model weights. 

The style classifier, performing binary classification, maps the rephrased text $y$ to a label $a\in \{0,1\}$. Since the classifier plays as a follower in the Stackelberg game, one can think of the label as the follower's action or response toward the LLM's paraphrasing.  Denote the classifier by $g(y)$, its definition is given by 
\begin{equation}
    a=g(y)\triangleq\left\{\begin{array}{ll}
      1   & \text{if $y$ is non-toxic}, \\
      0   & \text{otherwise}. 
    \end{array}\right.
\end{equation}
The sequel also uses $f_\theta(y|x)$ and $g(a|y)$ to represent the dependencies among variables.  
\paragraph{Stackelberg game}
To prepare the reader for the later introduced Stackelberg detoxification game, we use the notations above to review some basics. Beginning a payoff-relevant state $x$ (which affects the players' rewards/costs), the leader determines its parameterized policy $f_\theta$ that yields an action $y=f_\theta(x)$ upon receiving the state. Once observing the leader's action $y$, the follower responds by choosing action $a$. The follower's objective is to find a response policy $a=g(y)$ such that its reward $v(x,y,a)$ is maximized. The follower's best response given $x, y$ is $a^*=\arg\max_{a} v(x,y,a)\triangleq \operatorname{BR}(f_\theta(x))$. Aware of the follower's objective and best response, the leader aims to seek the optimal policy that maximizes its rewards (minimizes the cost) by soliciting the desired follower's action: $\max_\theta (x,a^*)=u(x,\operatorname{BR}(f_\theta(x)))$.     

\paragraph{Direct preference optimization}

DPO comes as a direct alternative to Reinforcement Learning from Human Feedback (RLHF), which consists of two main parts: the reward model $r_{\phi}$ and the target language model $\pi_{\theta}$. The reward model is optimized by the following loss function: $\mathcal{L}_\phi = -\mathbb{E}_{(x,y_w,y_l) \sim D} 
\left[ \log \sigma(r_{\phi}(x, y_w) - r_{\phi}(x, y_l)) \right]$, where $D$ is the dataset consisting of preference pairs $(x, y_w, y_l)$—where $x$ is an input, $y_w$, and $y_l$ are the preferred and dispreferred outputs. The $\sigma$ function here is sigmoid function. With a decent reward model $r_{\phi}$, we can then fine-tune the target language model $\pi_{\theta}$ by maximizing $
\mathbb{E}_{x \in D, y \in \pi_\theta} \left[ r_\phi(x, y) \right] - \beta D_{KL}(\pi_\theta(y|x) \| \pi_{\text{ref}}(y|x))$.

DPO has been gaining much popularity since it does not require a reward model and directly fine-tunes the LLM using preference pairs. The DPO loss is given by
\begin{align*}
    &\mathcal{L}_{DPO}(f_\theta; f_{\text{ref}}) = -\mathbb{E}_{(x, y_w, y_l) \sim D} \\
    &\left[ \log \sigma \left( \beta \log \frac{f_\theta(y_w \mid x)}{f_{\text{ref}}(y_w \mid x)} - \beta \log \frac{f_\theta(y_l \mid x)}{f_{\text{ref}}(y_l \mid x)} \right) \right].
\end{align*}

To understand how DPO works, one can inspect the gradient of DPO loss, which is given below. 
\begin{align*}
    &\nabla_\theta \mathcal{L}(\theta) \\
    &= - \mathbb{E}_{(x, y_w, y_l) \sim D} \bigg[ \underbrace{\beta \sigma (\hat{r}(y_l, x)  - \hat{r}(y_w, x) )}_{\mathrm{I}} \\
    & \left[ \underbrace{\nabla_\theta \log f_\theta (y_w \mid x)}_{\mathrm{II}} \right. \left. - \underbrace{\nabla_\theta \log f_\theta (y_l\mid x)}_{\mathrm{III}}\right] \bigg]
\end{align*}
The first term in the gradient increases when the reward estimation is wrong, so it can help LLM converge quickly. The second term can increase the likelihood of $y_w$ (the preferred text) and decrease the likelihood of $y_l$ (the dispreferred text).

\section{Stackelberg Learning from Response}


\subsection{Detoxification as  a Stackelberg game}
\label{subsec:stackelberg}
Consider the text detoxification task where the LLM seeks to set the model weights $\theta$ such that the paraphrased output $y\sim f_\theta(\cdot|x)$ is non-toxic. In other words, it aims to generate $y$ to pass the style classifier's screening, i.e., the desired classification outcome is $a=1$. 

We define the LLM's cost function as $u:D\times A\rightarrow \{0,1\}$ to be defined later in \Cref{sec:sro}, which encourages the LLM to generate successful paraphrases that pass the screening while penalizing the failed attempts.   Since the style classifier's objective is to report the underlying style accurately, one can define the style classifier's cost function as the classification loss. This work considers an SVM-based classifier and uses hinge loss \cite{NIPS1998_a14ac55a}. As discussed in the preliminary, we can treat the trained SVM as the best response mapping. 

The interaction between the LLM and SVM unfolds as below. First, the LLM takes as an input a text $x$ randomly sampled from the dataset $D$. Then, the LLM determines a paraphrase $y= f_\theta(x)$, which is then sent to the SVM. Finally, the SVM best responds with a classification outcome $a$, which decides the LLM's utility. The equilibrium problem for the LLM, the leader, is given by 
\begin{equation}
\label{eq:stackelberg}
    \min_{\theta\in \Theta} \sum_{x\in D} u(x,a^*) \quad \text{s.t. } a^*= \operatorname{BR}(f_\theta(x)),
\end{equation}
where $a^*$ simply corresponds to the classification output. Naturally, the solution to \eqref{eq:stackelberg} returns the optimal LLM model for detoxification. We introduce Stackelberg response optimization, adapted from DPO, to approximately solve for the optimal solution.     

\subsection{Stackelberg Response Optimization}
\label{sec:sro}
Given a preference pair $(y_w, y_l)$, DPO aims to solve the following minimization problem 
\begin{align*}
    & \min_{\theta \in \Theta} -\mathbb{E}_{(x, y_w, y_l)\sim D} \left[ \log \sigma \left( h^\beta_\theta(y_w|x) - h^\beta_\theta(y_l|x) \right)    \right]
\end{align*}
where we define 
    $h^\beta_\theta(y|x) \triangleq \beta \log \frac{f_\theta(y \mid x)}{f_{\text{ref}}(y \mid x)}$. In the detoxification task, whether preference pairs are available depends on the screening outcome. If the paraphrase $y$ passes the screening, then $(y,x)$ constitutes a preference pair, which can be handled by DPO. However, if $y$ fails the screening, then both $y$ and $x$ are dispreferred without a clear preference relation.   

We propose the following objective function to the LLM leader, which is a hybrid of two kinds of loss functions corresponding to cases of passing or failing the screening, respectively. \begin{align*}
    & \min_{\theta\in \Theta} \mathbb{E}_{x\sim D} u(x,a^*)\triangleq \mathcal{L}(\theta) \\
    & = -\mathbb{E}_{x\sim D} \left[ \log \sigma \left( a^* (h^\beta_\theta(y|x) - h^\beta_\theta(x|x))\right.\right. \\ 
    &\left.\left. +(1 - a^*) (-h^\beta_\theta(y|x)\right)  \right].
\end{align*}

\paragraph{Pass}

When the paraphrase passes the screening, i.e., $a = 1$, we adopt the standard DPO loss: 
\begin{align*}
    \mathcal{L}(\theta) = -\mathbb{E}_{x \sim D} \left[ \log \sigma \left( h^\beta_\theta(y|x) - h^\beta_\theta(x|x) \right) \right]
\end{align*}
We simply substitute $y_l$ with $x$ -- the input of the model. In our setup, the model will have a preference for the text ($y$) that passes the screening over the toxic input ($x$).

The gradient of the loss is
\begin{align*}
    &\nabla_\theta \mathcal{L}(\theta) = - \mathbb{E}_{x \sim D} \bigg[ \beta \sigma \left(h^\beta_\theta(x|x) - h^\beta_\theta(y|x) \right) \\
    &\left[ \nabla_\theta \log f_\theta (y \mid x) - \nabla_\theta \log f_\theta (x\mid x)\right] \bigg]
\end{align*}
Following the same argument in \cite{rafailov2024directpreferenceoptimizationlanguage}, we have $\hat{r}(x, y) = \beta \log \frac{f_\theta(y \mid x)}{f_{\text{ref}}(y \mid x)}$ is the implicit reward function defined by the LLM and the reference model. Then, by substituting $h^\beta_\theta(x|x)$ with $\hat{r}(x, x)$ and substituting $h^\beta_\theta(y|x)$ with $\hat{r}(y, x)$, the gradient of pass loss could be written as:
\begin{align*}
    &\nabla_\theta \mathcal{L}(\theta) \\
    &= - \mathbb{E}_{x \sim D} \bigg[ \underbrace{\beta \sigma (\hat{r}(x, x)  - \hat{r}(y, x) )}_{\mathrm{I}} \\
    & [ \underbrace{\nabla_\theta \log f_\theta (y \mid x)}_{\mathrm{II}}  - \underbrace{\nabla_\theta \log f_\theta (x\mid x)}_{\mathrm{III}}] \bigg]
\end{align*}
The first term in the gradient increases when the reward estimation is wrong (i.e., the reward difference is supposed to be negative). It serves as a coefficient that adjusts the gradient's magnitude so it can help LLM adapt quickly: when the reward estimation is wrong, the magnitude of the gradient is larger, and the LLM model weights change significantly. The second term determines the gradient direction to increase the likelihood of $y$ (the successful paraphrase) and decrease the likelihood of $x$ (the original input text). In summary, the gradient increases the likelihood of generating successful paraphrases.


\paragraph{Fail}
For those failed paraphrases, we apply the following loss to the model, which we call partial loss since it only keeps a part of the DPO loss. 
\begin{align*}
    \mathcal{L}(\theta)= -\mathbb{E}_{x \sim D} \left[ \log \sigma \left(- h^\beta_\theta(y|x) \right) \right].
\end{align*}

 Intuitively speaking, the partial loss does not indicate anything to prefer but only to disprefer. Thus, the loss leads the model to have a dispreference for the output ($y$). Below is the gradient of the partial loss.
\begin{align*}
- \mathbb{E}_{x \sim D} \bigg[- \beta \sigma \left(h^\beta_\theta(y|x) \right) \nabla_\theta \log f_\theta (y \mid x)   \bigg]
\end{align*}

Similarly, we can substitute $h^\beta_\theta(y|x)$ with $\hat{r}(y, x)$ and obtain the following. 
\begin{align*}
    &\nabla_\theta \mathcal{L}(\theta) \\
    &= - \mathbb{E}_{x \sim D} \bigg[- \underbrace{\beta \sigma (\hat{r}(y, x))}_{\mathrm{I}} 
     \underbrace{\nabla_\theta \log f_\theta (y \mid x)}_{\mathrm{II}} \bigg]
\end{align*}
In this gradient, the first term serves as the coefficient adjusting the gradient's magnitude. If the reward estimation is wrong (i.e., $\hat{r}(y, x)$ is large), then the gradient update significantly changes the model weights. The second term determines the gradient direction to decrease the likelihood of generating $y$. 

\section{Experiment}

\subsection{Toxicity Screener}
Our style classification model was developed using a dataset from the Jigsaw Toxic Comment Classification Challenge \cite{jigsaw-toxic-comment-classification-challenge}, which contains over 120,000 non-toxic sentences and 14,000 toxic sentences. Following the data splitting in \cite{logacheva-etal-2022-paradetox}, we allocate 95\% of the dataset for training and reserve 5\% for testing. We employ TF-IDF for sentence tokenization and SVM for classification to form the response. Ultimately, the screening model achieved an accuracy of 98.6\% on the test data.



\subsection{SRO Implementation}

We fine-tune the pre-trained paraphrase T5 model using the toxic sentences from the Jigsaw Toxic dataset \cite{jigsaw-toxic-comment-classification-challenge}.
The training process includes 15 epochs, with the batch size being 16. The learning rate is $2\times 10^{-5}$, and weight decay is set to 0.01. The hyper-parameter $\beta$ in SRP is 0.01.


\subsection{Competing Models}
We compare our SRO models with other style transfer models and common baselines in the literature \cite{dale-etal-2021-text}. 
\begin{itemize}
    \item \textbf{Duplicate} (baseline) – copy of the input.
    \item \textbf{Delete} (baseline) – deletion of swear words.
    \item \textbf{BART-zero-shot}  (baseline) – BART model with no additional training.
    \item \textbf{Mask\&Infill} \cite{ijcai2019p732} – BERT-based pointwise editing model.
    \item Delete-Retrieve-Generate models \cite{li-etal-2018-delete}: \textbf{DRG-Template} (replacement of toxic words with similar neutral words) and \textbf{DRGRetrieve} (retrieval of non-toxic sentences with a similar sense) varieties.
    \item \textbf{DLSM} \cite{He2020A} encoder-decoder model that uses amortized variational inference,
    \item \textbf{SST} \cite{lee-2020-stable} – encoder-decoder model with the cross-entropy of a pre-trained style classifier as an additional discriminative loss.
    \item \textbf{CondBERT}\cite{dale-etal-2021-text} – BERT-based model with extra style and content control.
    \item \textbf{ParaGeDi}\cite{dale-etal-2021-text} – a model which enhances a paraphraser with style-informed LMs which re-weigh its output.

\end{itemize}

\subsection{Metrics}

 We apply three third-party models and LLM to evaluate our model, which is customary in many style transfer works \cite{logacheva-etal-2022-paradetox, dale-etal-2021-text}. Namely, we evaluate:

\begin{itemize}
    \item style accuracy (STA) – the percentage of nontoxic outputs identified by a style classifier trained on the Jigsaw data. Note that such a testing classifier differs from the screener in that this classifier is built on a RoBERTa model and outputs a number ranging from 0 to 1, indicating toxicity likelihood.
    \item content preservation (SIM) – the cosine similarity between the embeddings of the input text and the paraphrase's embeddings \cite{wieting-etal-2019-beyond}. The embeddings are computed by an encoder that is trained on the ParaNMT corpus \cite{wieting2018paranmt50mpushinglimitsparaphrastic}. 
    
    \item fluency (FL) – the percentage of fluent sentences identified by a RoBERTa-based classifier of linguistic acceptability trained on the CoLA dataset\cite{10.1162/tacl_a_00290}.
\end{itemize}

Then, the \textbf{J} matric is computed by multiplying the three individual matrics: $\textbf{J}=\text{STA}\times \text{SIM}\times \text{FL}$. 

\subsection{Results}

\paragraph{Evaluation} Table \ref{tab:eval_result_loss} summarizes the performance of our proposed SRO, baselines, and prior works. Our model was trained on the Jigsaw Toxic Comment Classification Challenge dataset and it outperforms other models in terms of the \textbf{J} metric. The baseline models (Delete and Duplicate) and the BART-zero-shot model return low \textbf{J} scores due to the following reasons. The \textbf{Delete} model just deletes the toxic words in the sentence, leading to very low FL(fluency) scores. The \textbf{Duplicate} model has a very low STA score because it doesn't introduce any modification to the input. The \textbf{BART-zero-shot model}, as a pre-trained model trained on a large corpus of text, can only handle tasks based solely on its prior knowledge and the instructions given in the prompt. Thus, the BART-zero-shot model cannot perform the detoxification task, yielding a low STA score. 

Now we turn to other advanced competing models. \textbf{DRG-Template} and \textbf{DRG-Retrieve} models enjoy high STA scores because the first step of these models is to delete the tokens with toxicity, making the paraphrase pass the screening easily. However, the \textbf{Template} returns low FL as it just replaces the toxicity words, and the \textbf{Retrieve} gives low SIM because it retrieves a similar sentence from the training dataset with a non-toxic attribute, and the similarity in \textbf{Retrieve} is measured in Euclidean distance on the embedding of two sentences. However, a small Euclidean distance does not imply content preservation. \textbf{Mask\&Infill} model masks the toxic markers and infills them with similar non-toxic tokens. Such a practice maintains the basic structure of the input text, leading to a high SIM score. However, the infilling may introduce words that render the whole text incomprehensible and result in a low FL score. The \textbf{DLSM} and \textbf{SST} models perform full re-generation of text (as opposed to word-wise corrections). More importantly, their decoders are trained from scratch on a relatively small dataset, leading to low fluency scores. The \textbf{ParaGeDi} model applies a re-weight idea on logits to control the attribute of output, which means it pays little attention to the context. Therefore, this model gives a relatively low FL score. This problem also happens to \textbf{CondBERT}, which also uses re-weight on logits-wise, so the FL of generation text becomes relatively low.

In contrast to the models above that perform word/logit-wise corrections, our SRO model, fine-tuned on a text-paraphrase LLM, is more concerned with the context, yielding a high FL score. As shown in Table~\ref{tab:eval_result_loss}, our SRO achieves the highest FL scores while retaining satisfying scores on STA and SIM. Of particular note, SRO outperforms all the advanced competing models in terms of SIM.   

\begin{table}[h!]
  \centering
  \caption{Comparison of models on various metrics and the ablation tests on the loss design.}
  \small
  \label{tab:eval_result_loss} 
  \begin{tabular}{l|c|c|c||c}
    \midrule
    Model  & STA & SIM & FL & J \\
    \midrule
    Human reference  & 0.96 & 0.77 & 0.88 & 0.66 \\
    \midrule
    SRO   & 0.86 & 0.84 & 0.92 & $\mathbf{0.66}$\\
    PA-SRO   & 0.34 & 0.67 & 0.88 & 0.20\\
    RS-SRO &  0.71 & 0.78 & $\mathbf{0.93}$ & 0.51\\

    Delete  & 0.81 & 0.93 & 0.64 & 0.46 \\
    Duplicate  & 0.02 & 1.00 & 0.91 & 0.02 \\
    DRG-Template  & 0.90 & 0.82 & 0.69 & 0.51 \\
    BART-zero-shot  & 0.01 & 0.99 & 0.92 & 0.01 \\
    Mask\&Infill  & 0.91 & 0.82 & 0.63 & 0.48 \\
    CondBERT  & 0.98 & 0.77 & 0.82 & 0.62 \\
    SST & 0.86 & 0.57 & 0.19 & 0.10 \\
    ParaGeDi  & 0.99 & 0.71 & 0.88 & 0.62 \\
    DLSM  & 0.76 & 0.76 & 0.52 & 0.25 \\
    DRG-Retrieve  & 0.97 & 0.36 & 0.86 & 0.31 \\ 
    \bottomrule
  \end{tabular}
\end{table}

Table \ref{tab:example} shows examples of different models' output, and we can see that our model's results look more tactful and retain the original meaning.

\begin{table*}[ht]
  \centering
  \caption{Examples of detoxification by different models.}
  \label{tab:example}
  \begin{tabular}{@{}p{3cm}p{12cm}@{}}
    \toprule
    \textbf{Input} & \textbf{Output} \\ 
    \midrule
    Input & no room for those stupid antics. \\
    SRO & No room for those bad exploits. \\
    ParaGeDi & There’s no space for all those adversities. \\
    CondBERT & no room for those . . . those bizarre exploits. \\
    Mask\&Infill & no room for those compelling antics. \\
    DRG-TemplateBased & no room for those conventions antics. \\
    \midrule
    Input & zoological conspiracy theories (Arab-Israeli conflict), by contrast, sucks horribly. \\
    Ours & Zooological conspiracy theories (Arab-Israeli conflict), by contrast, is not good. \\
    ParaGeDi & By contrast, the zoological theories of conspiracy (the Arab-Israeli conflict) suffer horribly. \\
    CondBERT & zoological conspiracy theories (Arab-Israeli conflict), by contrast, worked out badly. \\
    Mask\&Infill & zoological conspiracy theories (Arab-Israeli conflict), by contrast, ends horribly. \\
    DRG-TemplateBased & zoological conspiracy theories (Arab-Israeli conflict), by contrast, instrumental horribly. \\
    \midrule
    Input & this whole article is bullshit. \\
    Ours & This whole article is not that good.\\
    ParaGeDi & This article’s not a good deal. \\
    CondBERT & this whole article is considered incomplete. \\
    Mask\&Infill & this whole article is compelling. \\
    DRG-TemplateBased & this whole article is substantial. \\
    \bottomrule
  \end{tabular}
\end{table*}

\subsection{Ablation study on SRO loss} 
\paragraph{RS-SRO}
The difference between SRO and DPO lies in the use of paraphrased text that fails the screening, for which no preference exists between the input and the output. To see how the failed attempts contribute to the success of SRO, we conduct an ablation study where SRO discards the failed outputs and only performs DPO on those successfully paraphrased instances. We refer to such a practice of SRO as response-selective SRO (RS-SRO) since the leader (LLM) only considers positive responses $a^*=1$ while ignoring those negative ones. The loss function of RS-SRO is given by $\mathcal{L}(\theta)= -\mathbb{E}_{(x, y) \sim D} [ \log \sigma ( h^\beta_\theta(y|x) - h^\beta_\theta(x|x) ) ]\mathds{1}_{\{a=1\}}$.
From Table \ref{tab:eval_result_loss}, we can see that the performance of the RS-SRO model decreases by a small margin compared to the SRO model, which proves the use of the fail loss we apply to the SRO model contributes to the SRO's success.

\paragraph{PA-SRO}

The above experiment demonstrates the role played by negative feedback without creating complete preference. Then, one naturally wonders if SRO's treatment of failed paraphrases would also carry over to the successful ones. Even though the pair of the input and output reveals a preference relation, it is unclear whether such a relation is needed in SRO. Using terms in the Stackelberg model, the question is whether positive feedback with complete preference is necessary. To answer this question, we modify the SRO loss when $a^*=1$ and refer to the resulting fine-tuning as preference-agnostic SRO (PA-SRO). Now for the PA-SRO model, the pass doesn't contain the $h^\beta_\theta(x, x)$ part in the pass loss. The loss function of PA-SRO is given by $\mathcal{L}(\theta)=-\mathbb{E}_{(x, y) \sim D} [ \log \sigma ( a^*h^\beta_\theta(y|x)-(1 - a^*) h^\beta_\theta(y|x) ]$. From the results in Table \ref{tab:eval_result_loss}, we can find the PA-SRO's \textbf{J} score is significantly smaller than those of SRO and RS-SRO. In our loss design, the partial loss (fail) is important though, the DPO-type loss (pass) is much more important as it creates a preference pair for the model to learn a direct preference instead of giving an indirect instruction for the model to learn.

\paragraph{Robustness to Misclassification} 
The style screener plays an instrumental part in the proposed SRO, providing feedback to the LLM's paraphrase. To investigate the screener's robustness against possible misclassification in style screening, we introduce various levels of misclassification to the screener and collect detoxification metrics when the LLM learns from possibly inaccurate responses.  

We consider the following levels of misclassification: 10\%, 20\%, 50\%, and 70\%. For a given misclassification level, such as 10\%, the working of the screener proceeds as below. Given a paraphrased text from the LLM, the screener employs the trained SVM to label the text with 90\% probability while classifying the text as non-toxic regardless of the true style with 10\%. In other words, the introduced misclassification serves as a relaxation of the screening process, creating more preference pairs with possibly inaccurate preference relations. In this case, an originally unsuccessful paraphrase, together with the toxic input, creates a seemingly complete preference pair to be fed to SRO. By conducting such experiments, we can determine how vulnerable the LLM is to misleading preference relations. From Table \ref{tab:eval_result_screener}, we can see a huge gap between \textbf{J} of the SRO model and SRO under 10\% relaxation (denoted by R-10\%). The \textbf{J} also slightly decreases while the confidence decreases. We can conclude that the credibility of the screener is extremely important for the performance of SRO. In our understanding, the misleading preference pairs greatly hinder the SRO model from converging and give the SRO model wrong information or preferences to learn. This greatly reduces learning efficiency because making an error requires more effort to adjust it.

\begin{table}[h!]
  \centering
  \caption{Ablation test on the screener}
  \small
  \label{tab:eval_result_screener} 
  \begin{tabular}{l|c|c|c||c}
    \midrule
    Model  & STA & SIM & FL & J \\
    \midrule
    Human reference  & 0.96 & 0.77 & 0.88 & 0.66 \\
    \midrule
    SRO   & 0.86 & 0.84 & 0.92 & $\mathbf{0.66}$\\
    R-10\%  & 0.53 & 0.78 & 0.92& 0.38\\
    R-20\%  & 0.48 & 0.78 & 0.92 & 0.34\\
    R-50\%  & 0.42 & 0.74 & 0.88 & 0.27\\
    R-70\%  & 0.40 & 0.77 & 0.87 & 0.26\\
    K-10\%  & 0.50 & 0.67 & 0.92 & 0.30 \\
    K-20\%  & 0.39 & 0.69 & 0.90 & 0.24 \\
    K-50\%  & 0.35 & 0.66 & 0.90 & 0.20 \\
    K-70\%  & 0.33 & 0.67 & 0.91 & 0.20 \\
    \bottomrule
  \end{tabular}
\end{table}

In contrast to relaxation, another type of misclassification is overkill; that is, the screener classifies the paraphrase as toxic, whatever the true style is. In this case, the overkill greatly reduces the portion of preference pairs in the response. Consequently, the partial loss is more frequently employed in the fine-tuning process. Similar to the previous setup, we introduce 10\%, 20\%, 50\%, and 70\% overkill to the screener to investigate the LLM's robustness against such misclassification. From the result shown in Table \ref{tab:eval_result_screener}, we can see that the performance under the overkill screener decreases as the overkill rate increases. Besides, as we compare the relaxation case and overkill case at the same rate, we can conclude that the complete preference pairs when passing the screening are more helpful than the partial loss in the failed cases. 

\section{Conclusion}
In this paper, we provide a new Stackelberg-game perspective on text detoxification using LLM on non-parallel data, where the LLM acts as the leader, aiming to determine its model weights such that the paraphrased texts will pass the toxicity screener. Such a Stackelberg formulation spares one from collecting human preference labeling and enables the LLM to learn directly from the screener's response. We propose Stackelberg response optimization (SRO) to solve the optimal leader's policy for LLM detoxification. Experimental results indicate that our SRO achieves superior overall performance over other competing models and matches human performance. Of particular note is that SRO's success depends largely on the screener's accurate feedback; a slight perturbation in the response leads to significant performance degradation.

\bibliography{custom}
\bibliographystyle{acl_natbib}

\end{document}